\documentclass[letterpaper, 10 pt, conference]{ieeeconf}  
\usepackage{cleveref}
\usepackage{mathrsfs}
\makeatletter
\let\NAT@parse\undefined

\IEEEoverridecommandlockouts                              

\usepackage{amsmath,amsfonts,amssymb}
\usepackage{url}
\usepackage{graphicx}
\usepackage{float}
\usepackage{multirow}
\usepackage{tikz}
\usepackage{pgf}
\usepackage{placeins}
\usepackage{subcaption}
\usepackage{comment}
\usepackage{amsmath}
\usepackage{placeins}
\usepackage{latexsym}
\usepackage{pifont}
\usepackage{pgfplots,filecontents}
\usepackage{gensymb}
\usepackage{epstopdf}
\usepackage{siunitx}
\pgfplotsset{compat=newest}

\usepackage[symbol]{footmisc}

\usetikzlibrary{intersections}

\DeclareGraphicsExtensions{.png,.jpg,.jpeg,.eps,.gif}
\title{\LARGE \bf
Design, Development and Experimental Realization of a Quadrupedal Research Platform: Stoch
}

\author{Dhaivat Dholakiya, Shounak Bhattacharya, Ajay Gunalan, Abhik Singla, \\  Shalabh Bhatnagar, Bharadwaj Amrutur, Ashitava Ghosal and Shishir Kolathaya
\thanks{This work is supported by the Robert Bosch Center for Cyber Physical Systems, Bangalore, India}
\thanks{Dhaivat Dholakiya*, Shounak Bhattacharya, Ajay Gunalan, Abhik Singla are with the Robert Bosch Centre for Cyber-Physical Systems, IISc, Bangalore, India. 
        {E-mail: \tt\small \{dhaivatd, shounakb, ajayg, abhiksingla\}@iisc.ac.in}.}%
\thanks{Ashitava Ghosal is with the Faculty of Mechanical Engineering, Bharadwaj Amrutur is with the Faculty of Electrical \& Computer Engineering, Shalabh Bhatnagar is with Computer Science \& Automation, and Shishir Kolathaya is an INSPIRE Faculty Fellow at the Robert Bosch Center for Cyber Physical Systems, IISc, Bengaluru, India
{\tt\small \{shalabh,amrutur,asitava,shishirk\}@iisc.ac.in}}%
}

\begin{document}

\maketitle
\thispagestyle{empty}
\pagestyle{empty}

\begin{abstract}
In this paper, we present a complete description of the hardware design and control architecture of our custom built quadruped robot, called the \textit{Stoch}.
Our goal is to realize a robust, modular, and a reliable quadrupedal platform, using which various locomotion behaviors are explored. 
This platform enables us to explore different research problems in legged locomotion, which use both traditional and learning based techniques.
We discuss the merits and limitations of the platform in terms of exploitation of available behaviours, fast rapid prototyping, reproduction and repair.
Towards the end, we will demonstrate trotting, bounding behaviors, and preliminary results in turning. In addition, we will also show various gait transitions i.e., trot-to-turn and trot-to-bound behaviors.

\end{abstract}


\section{Introduction}

In the current state of research in robot locomotion, a beginner usually faces a stiff learning curve to design and fabricate a robust and a reliable quadrupedal walking platform. Crossing this barrier is not only limited by the individual's capacity but also by the requirement of resources. Therefore, a robot which is simpler, modular, affordable, easily repairable, and yet facilitating complex behaviors is largely preferred.



\begin{figure}[t!]
\centering
\includegraphics[width=\linewidth]{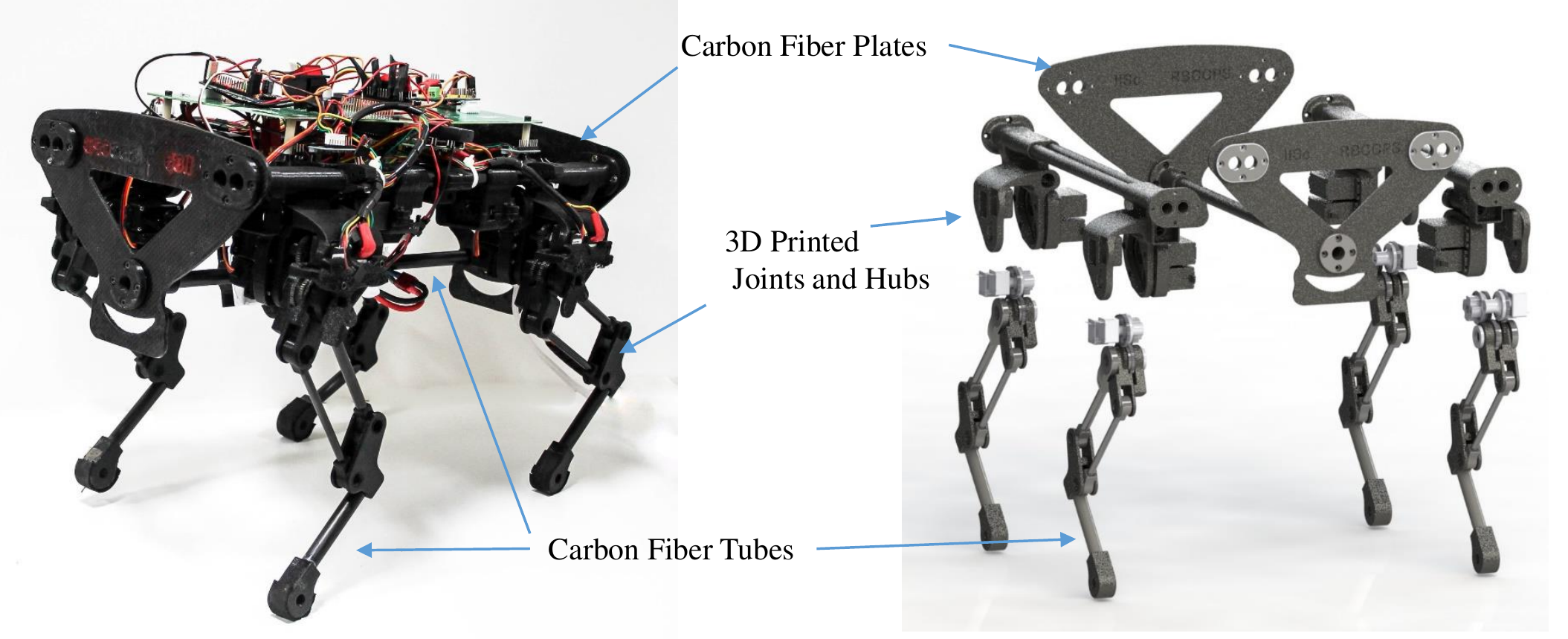}
  \caption{Left image shows the physical hardware of Stoch, and the right image shows its rendered exploded view. }
  \label{fig:overall_pic}
\end{figure}

With a view toward realizing state of the art walking controllers in a  low-cost walking platform, we developed a quadrupedal robot called the \textit{Stoch}. This manuscript primarily serves as a detailed description of the hardware and software design process and construction of this platform (see \Cref{fig:overall_pic}), which complements the results that were shown in \cite{singla2018realizing}. 
This hardware features modular central body and a five bar co-axial leg design. Which are fabricated using easily available off the shelf materials like carbon-fiber tubes and custom 3D PLA parts. Carbon fiber tubes, which are strong and light, serve as the structural members and leg links. Similarly, the 3D printed parts, manufactured in-house, act as the joints. The software consists of a real-time trajectory generator, which are then passed as commands to the individual servo drives. The trajectory generator performs reasonably well at control frequencies on many off-the-shelf servos (50 Hz). 
In addition to realizing robust gaits in the real hardware, the central pattern generator (CPG) based trajectory generator \cite{sprowitz2014kinematic, sprowitz2013towards, sprowitz2018oncilla} also enables 1. Realization of derived walking gaits like bound, walk, gallop, and turning in hardware with no additional training, and 2. Smooth transitions from one type of gait to another. This is mainly motivated by the various locomotion behaviors realized by \cite{ijspeert2008central,sprowitz2014kinematic} in their custom hardware. These various behaviors are then composed together and tele-operated by a remote control device.

The paper is organized as follows: \Cref{sec:design} details the design and hardware of Stoch. \Cref{elec-software-architecture} describes the software architecture, on-board compute platform, actuators, communication interfaces, and the sensory inputs.
\Cref{sec:control} contains a detailed  description of the control algorithm used. Finally \Cref{sec:results} describes the experimental results of Stoch demonstrating multiple types of gaits and the associated transitions.


\section{Related Work}\label{sec:related_work}


There have been multiple types of quadruped robots in varying sizes over the past ten years, namely,
MIT Cheetah  \cite{park2015variable}, Starl ETH \cite{hutter2012starleth}, Spot mini  \cite{SpotMini}, ANYmal \cite{hutter2016anymal}, LaikaGo \cite{Laikago}, Minitaur \cite{de2018vertical} etc. Most of these works have mainly focused on answering important questions on control and hardware design for walking, and have also demonstrated multiple types of functionalities both in simulation and experiments. Despite their impressive results, it is important to note that these robots cost to the tune of minimum \$30,000, are expensive to manufacture, and also require advanced sensing and actuation.


There are also quadrupeds that are less resource intensive such as Tekken \cite{kimura2004biologically}, Cheetah-cub \cite{sprowitz2013towards}, Onchila \cite{sprowitz2018oncilla}, Mini-cheetah \cite{bosworth2015super}, Aibo \cite{Aibo}, Rhex \cite{Rhex}, some of which are open sourced, but not easy to reproduce. 

\section{Mechanical Design}
\label{sec:design}

Stoch  is  a  quadrupedal  robot  
designed  and developed in-house at the Indian Institute of Science (IISc), Bangalore, India. 
The design philosophy behind Stoch is based on modularity, lightweight construction, ease of manufacturing, rapid repair and reproduction.
\begin{figure}
\centering
\includegraphics[width=\linewidth]{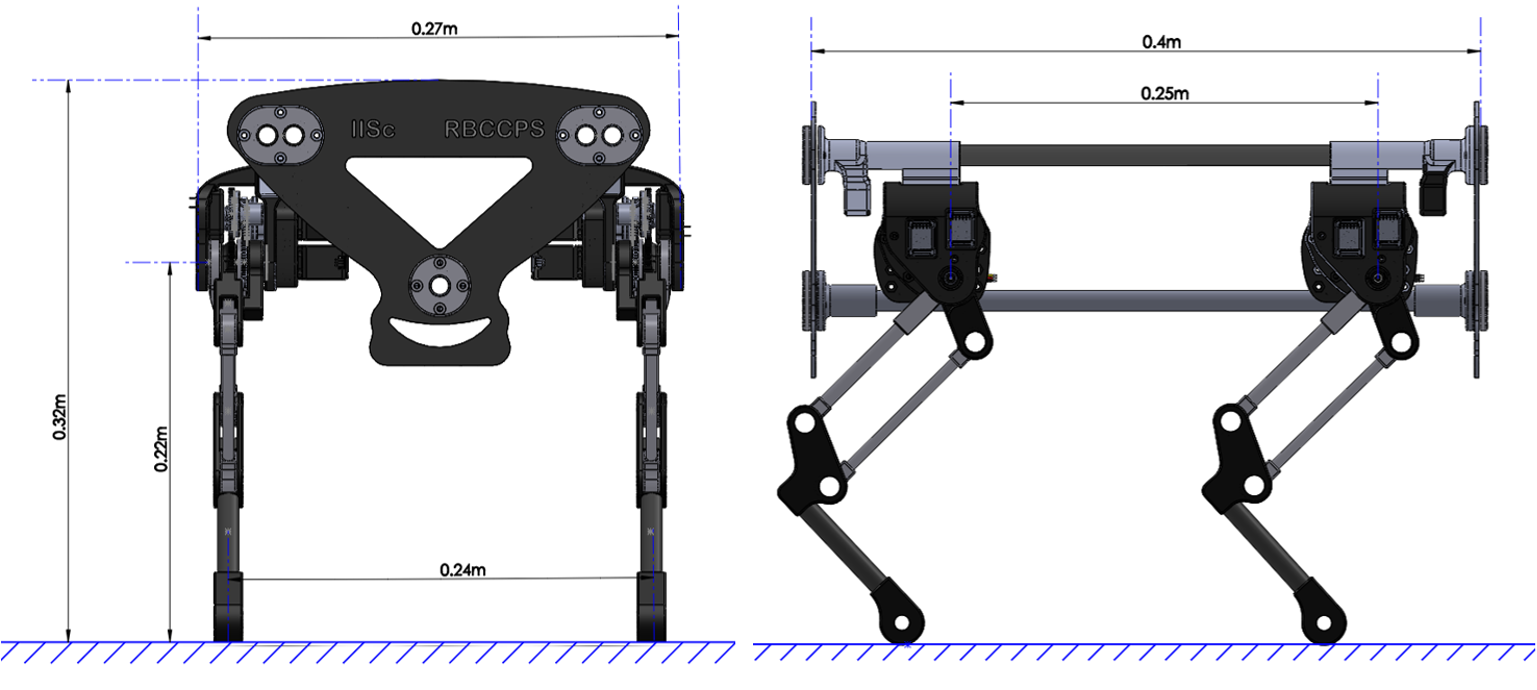}
  \caption{Physical robot dimensions of Stoch.}
  \label{fig:robo_dim}
\end{figure}

\begin{figure*}
\centering
\includegraphics[width=\linewidth]{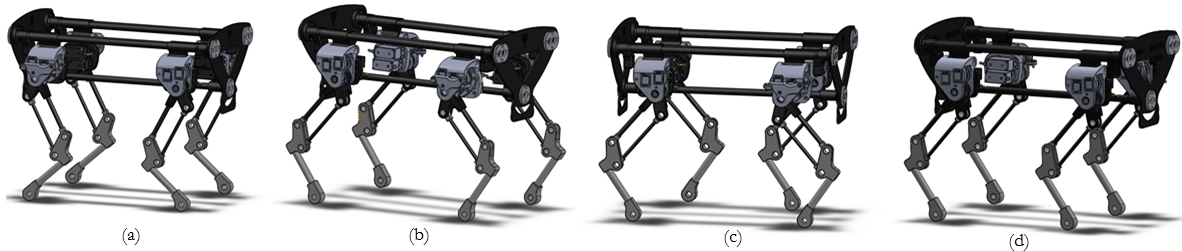}
  \caption{Modular Configurations of \textit{Stoch}: (a) X configuration, (b) O configuration, (c) XO configuration, (d) Natural quadruped configuration.}
  \label{fig:config}
\end{figure*}

The robot design can be understood as an assembly of a central body module and four leg modules.
\Cref{fig:overall_pic} shows an exploded view showing all of the modules of Stoch. The leg modules can be used in several types of morphologies as shown in \Cref{fig:config}.
The body consists of carbon fiber (CF) hollow tubes along its length as strong and lightweight structural members. The front and back CF plates along with 3D printed poly lactic acid (PLA) hubs form rigid central frame structure. The body also houses all electronic parts, cable routing and battery power supply. A salient feature of this design is that the leg modules slide and `sit' on CF tubes placed along the body length. Thus, the geometrical distance between front and hind legs can be modified by changing the CF tube lengths or the assembly positions of the leg modules. Similarly, the distance between the left and right legs, which is constrained by the CF plates, can easily be customized. 
This feature enables user to rapidly play and experiment with the basic geometry of legs and body module. 

In this section, we first discuss the overall geometry of the mechanical structure, and then focus on its individual components, i.e., body and leg design. Finally, we describe the actuator and sensor arrangement on the legs.

\subsection{Geometry of Stoch}
\label{sec:geometry}

The Stoch is designed equivalent to the size of a miniature Pinscher dog. There are four identical legs and eight actuators mounted in the robot, which are distributed equally. Each leg has a hip and a knee joint, and the actuators provide flexion/extension in each joint. As shown in \Cref{fig:robo_dim}, the Stoch is $0.4$ \si{m} long in total, with $0.25$ \si{m} between front and hind legs. It has a maximal hip height of $0.24$ \si{m} with legs at maximum stretch, a standing height of $0.22 $ \si{m}, and an overall height of $0.35 $ \si{m} from ground to the top of electronics and cables. The robot has $0.24 $ \si{m} of lateral spacing between left and right leg planes axes, and a maximum width of $0.27 $ \si{m}. The robot weighs around $3.1$ \si{kg} and $3.7$ \si{kg} with the battery-pack. The estimated static centre of mass (COM) of the given system (in standing configuration similar to \Cref{fig:robo_dim}) is situated $35$ \si{mm} below the hip axis.

\begin{table}[h!]
 \centering
 \begin{tabular}{l r}
 \hline
 \textbf{Parameter} & \textbf{Value}\\
 \hline
total leg length & 230 mm\\
leg segment length & 120 mm\\ 
min./max. hip joint & -45$^{\circ}$/ 45$^{\circ}$\\
min./max. knee joint & -70$^{\circ}$/ 70$^{\circ}$\\
max hip speed & 461$^{\circ}$/s \\
max knee speed & 461$^{\circ}$/s \\
max hip torque & 29 kg-cm\\ 
max knee torque & 35 kg-cm\\
motor power (servo) & 16 W \\

 \hline
 \end{tabular}
 \caption{Hardware specifications of leg. 
 }
 \label{tab:robot_leg}
 \end{table}
 
\subsection{Leg design}
\label{sec:Leg_design}

Every leg has two degrees of freedom, one each for hip and knee. Detailed view of the leg assembly is shown in \Cref{fig:leg}. The key specifications of leg and actuators  are summarized in \Cref{tab:robot_leg}.
The emphasis in the leg design was to keep the inertia of the moving segments minimal. To achieve this, all the actuators and transmissions are mounted to the main body. The leg mechanism is designed as a co-axial five bar linkage\cite{de2018vertical}. This is beneficial to ensure fast swing motions with minimum leg inertia.

To fabricate the mechanism, the linkages are formed with carbon fiber hollow tubes connected by 3D printed poly lactic acid (PLA) connectors. This arrangement enables us to modify the link lengths with minimal effort. To further reduce the swing leg inertia, we've used CF tubes as journal bearings and shaft at pivot joints instead of regular metal ball bearings. This drastically reduces the weight of the robot.

\subsection{Actuation and sensing arrangement in legs}
\label{sec:ActuationAndSensingArrangementinLegs}

\begin{figure*}[t!]
\centering
\includegraphics[width=0.8 \linewidth]{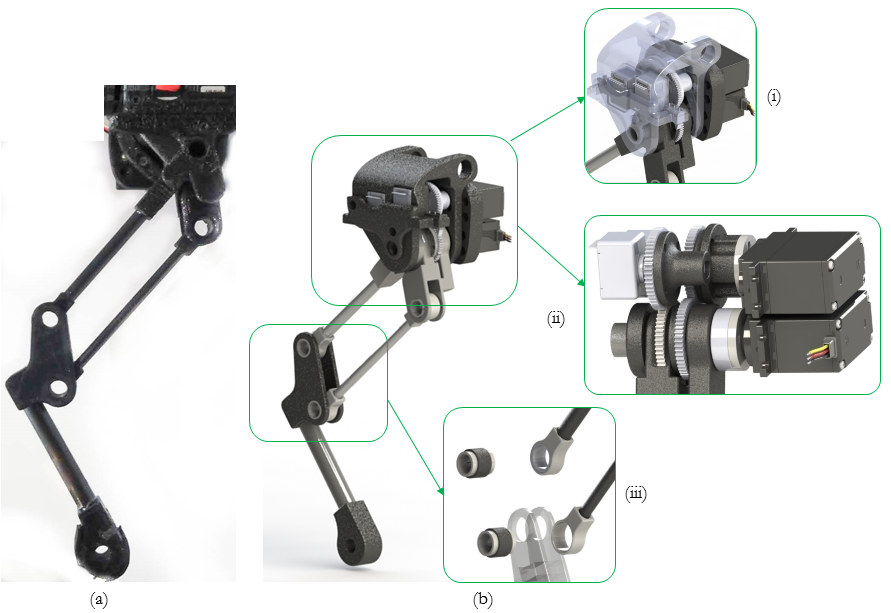}
\caption{(a) shows the physical leg assembly, (b) shows the rendered counterpart of the leg assembly. (b) has three parts: (i) sensor mounting and actuation hub, (ii) coupling and transmission of linkages with actuators and rotary encoders, and (iii) exploded view of knee joint and CF journals.}
\label{fig:leg}
\end{figure*}

\Cref{fig:leg} (i) and (ii) depicts the assembly arrangement of servo actuators to the leg linkages and rotary encoders. For the joint actuation we have used standard servo motors (Knee: Robokits $35$ \si{kg-cm} High Torque Servo
, Hip: Kondo $2350$HV Servo.
The knee actuators are directly coupled with the $3$D printed links, which in turn move the coupler links and eventually the knee joints of the legs. The joint angle sensor (rotary encoder) for knee is connected with the linkage via a laser cut steel gear. Similarly, the hip actuator transfers mechanical power to the hip link via a simple gear train (ratio $1:1$), whereas the hip encoder is directly coupled to monitor the hip servo angle via a $3$D printed part.

\hypertarget{software-architecture}{%
\section{Electronic System Architecture}\label{elec-software-architecture}}

\begin{figure}[h!]
\includegraphics[width = \linewidth]{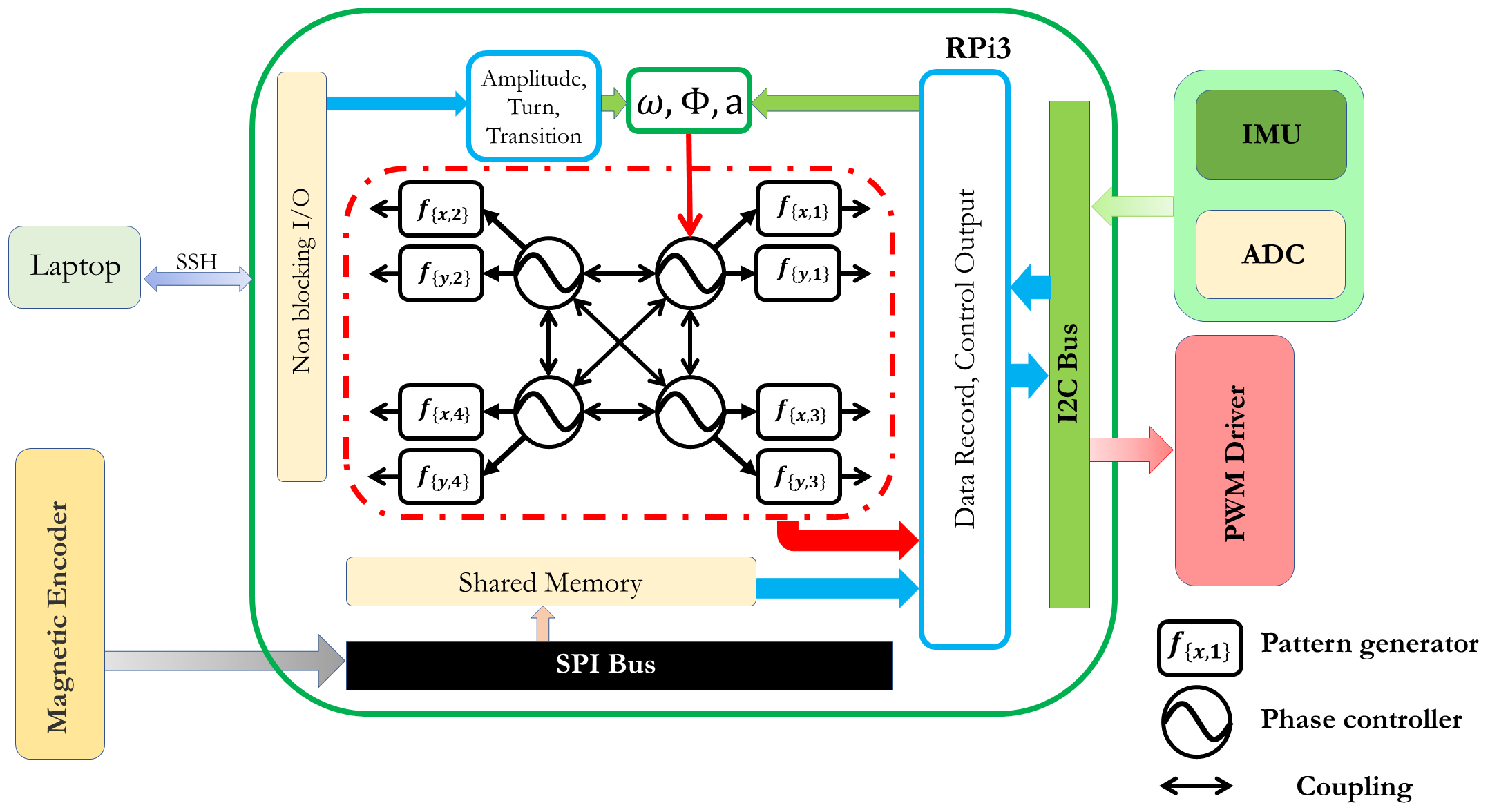}
\caption{Controller and tele-operation schematic. }
\label{fig:circuit_control}
\end{figure}

In this section, we describe the electronic system architecture 
in light of the objectives achieved in \textit{Stoch}. 
The platform achieved three main objectives using this architecture.
First, tele-operation of the robot, which is implemented via SSH over wireless communication (\Cref{fig:circuit_control}). 
Second, the low-level implementation of the walking controller
, which passes on the desired joint configurations as commands to the  joint actuators. 
And third, integrating sensory feedback with the walking controller. The last one is achieved by reading and  recording different sensors without experiencing significant delay. 

The encoders, servos and other parts selected for this work are readily available, off-the-shelf components. This enabled a quick adoption, rapid prototyping while minimizing the total cost. 

\begin{table}[h!]
 \centering
 \begin{tabular}{|l|r|}
 \hline
 \textbf{Equipment} & \textbf{Details}\\
 \hline
Joint encoder & Bourns EMS22a\\
IMU & MPU-$9150$\\
ADC & ADS1115\\
PWM driver & Adafruit PCA968\\
Servo & Kondo $2350$HV\\
 &  Robokits RKI-$1203$\\
 Computer & Raspberry Pi 3b\\
 \hline
 \end{tabular}
 \caption{Electronics hardware specifications. }
 \label{tab:robot_electronics}
 \end{table}
\subsection{Tele-operation}\label{sec:teleoperation}
The walking controller operates by changing internal parameters of the Central Pattern Generator (CPG), such as the frequency, phase difference and amplitude. 
These changes result in generation of different behaviors and motions.
In this robot, these parameter changes, are made during the process execution via SSH. 
We selected the keyboard as the input device for remote operation. 


\subsection{Motor actuation}\label{sec:actuator_control}
The actuators used in this robot are PWM enabled servo motors. To actuate these servos, we have used Adafruit PWM drivers
, connected with the central compute platform (Raspberry Pi3) over I2C Bus. 
In this robot, there are two sets of servos (see \cref{sec:ActuationAndSensingArrangementinLegs}). These servos operate at $11.1$ V and $7.4$ V respectively. 


\subsection{Sensory feedback}\label{sec:sensory_input}
The sensory inputs contain signals from joint encoders, inertial measurement unit and limit switches. 
The angular position of each motor shaft is measured by externally coupled absolute magnetic encoders. 
Each encoder is read sequentially by the RPi3 via a multi-duplex, SPI Bus. 
The low-level drivers 
are used to pass the values to the primary process 
in real-time using shared memory.
We used the limit switches for self-calibration and failure prevention at the joints. 
These switches are connected to an analog to digital converter on the I2C Bus.

The center of mass of the robot, is estimated by the inertial measurement unit  (IMU). 
The states observed by the IMU are three axis acceleration, three axis angular velocity etc.
Using sensor fusion, the IMU can also estimate the roll, pitch, yaw of the robot.
The IMU is placed close to the geometric center of the chassis and is connected to the I2C bus. 

\begin{figure*}[h!]
\centering
\includegraphics[width = 0.75\linewidth]{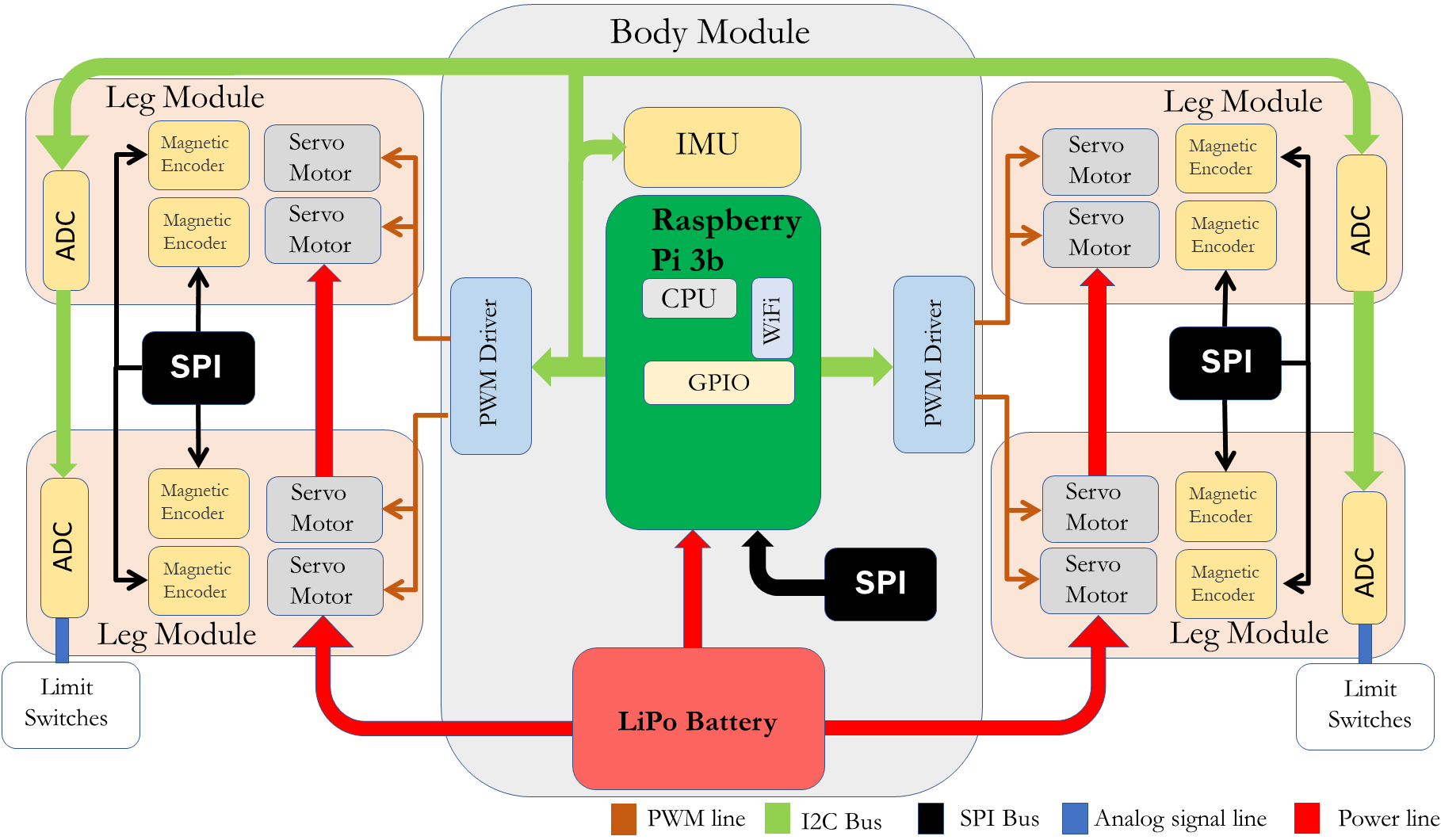}
\caption{Schematics of electronics and communication network.}
\label{fig:circuit}
\end{figure*}




\section{Walking Controller}\label{sec:control}
In this section, we describe the design of the control architecture of \textit{Stoch}. The controller is based on the use of nonlinear coupled oscillators, called as \textit{Central Pattern Generator} (CPG) \cite{ajallooeian2013general,sprowitz2018oncilla}. This approach uses a group of polynomials associated with the coupled oscillators to generate the desired patterns at the end point.
We will first describe the methods used to develop the rhythmic patterns using CPG, and then discuss inverse kinematics used to transform these patterns into desired joint angles.

\subsection{Central pattern generator} \label{sec:PatternGenerator}
Central pattern generators (CPGs) are neural networks capable of producing coordinated patterns of rhythmic activity without any rhythmic inputs from sensory feedback or from higher control centers \cite{ijspeert2008central}. 
In this work, the neural network has been replaced with nonlinear coupled differential equations. 
When a signal is provided  with  oscillation frequency ($\omega$), the CPG  begins to produce rhythmic signals of a predetermined pattern.
The oscillation frequency updates the phases ($\phi$) and the coupling ensures the phase offsets ($\Phi$) between the oscillators. This difference in phase offset results in the generation of different types of gaits.
The equations for the CPG are given as
\begin{align}
    & \dot{\omega} = \alpha_{\omega} (\omega_d - \omega) \label{eq:lowpassfilter}\\
    & \dot{\phi_i} = \omega + \sum_{j=1}^N K_{i,j}\sin \left(\phi_i - \phi_j + \Phi_i -\Phi_j \right)\\
    & \phi_{i} = mod(\phi_{i}, 2\pi)\\
    & \dot{\mathbf{X}} = \dot{\mathbf{\phi}} \mathbf{X}^{\prime}_{d} + \alpha_{\mathbf{X}}(\mathbf{X}_{d} - \mathbf{X}) + \Delta \label{eq:modifiedlowpass}
\end{align}

where, $\omega, \omega_d$ are the current and target oscillation frequency of the legs, in radian per second. $\phi_i$ is the phase of the $i^{th}$ internal pattern generator. 
$\phi$ is the vector containing the $\phi_i$. 
$\Phi_i$ is the phase difference between the $i^{th}$ leg and the first leg. $K_{i,j}$ are coupling constants.
\Cref{eq:lowpassfilter} and (\ref{eq:modifiedlowpass}) represent low pass filter in time domain \cite{ijspeert2008central}. 
These filters ensure there are no discontinuous changes in the variables. 
The low pass filter in  \Cref{eq:modifiedlowpass} \cite{ajallooeian2013general} is modified such that, periodic signal of any desired frequency can pass through but undesired signal of the high frequency can not. 
In \Cref{eq:modifiedlowpass}, $\mathbf{X}, \mathbf{X}_d$ are the current and the desired state values and $\mathbf{X}^{\prime}_{d} = \partial \mathbf{X}_{d} / \partial \mathbf{\phi}$.
The constants $\alpha_{\omega}, \alpha_{\mathbf{R}}$ are used to determine the corner frequency of these filters. 
$\Delta$ is the discrete or event based input provided by the user or the sensors on-board. 
In this work, the filters are used in places where a discrete signal is added by the user or by some feedback in response to an event. 

After the phase ($\phi_i$) is determined, any function described in form of $\phi_i$ can be used to create the endpoint behaviour of the $i^{th}$leg. We have
\begin{align}
    & \mathbf{X}_{d,i} = \mathbf{X}_d(\phi_i) = \left[x(\phi_i)~~ y(\phi_i)\right]\\
    & x(\phi_i) =\sum^n_{j=1}W_{x,j,i}f_{x,j}(\phi_i)
\end{align}

In the above equations, $x(\phi_i)$, $y(\phi_i)$ represent the desired end point position values of the leg from the hip joint of the $i^{th}$ leg.
This scheme can also be used to generate $r(\phi_i)$, $\theta(\phi_i)$ in polar co-ordinates which are used in the reinforcement learning framework \cite{singla2018realizing}.
The constant vectors, $W_{x,i,j}$, are the weights associated with the basis functions, $f_{x,j}$. Similarly we can generate $y(\phi_i)$ with weights and basis function. The patterns generated in Cartesian or in polar co-ordinates are then transformed into joint co-ordinates via inverse kinematics.

\subsection{Inverse kinematics}
\begin{figure}[h!]
\centering
\includegraphics[width = 0.85\linewidth]{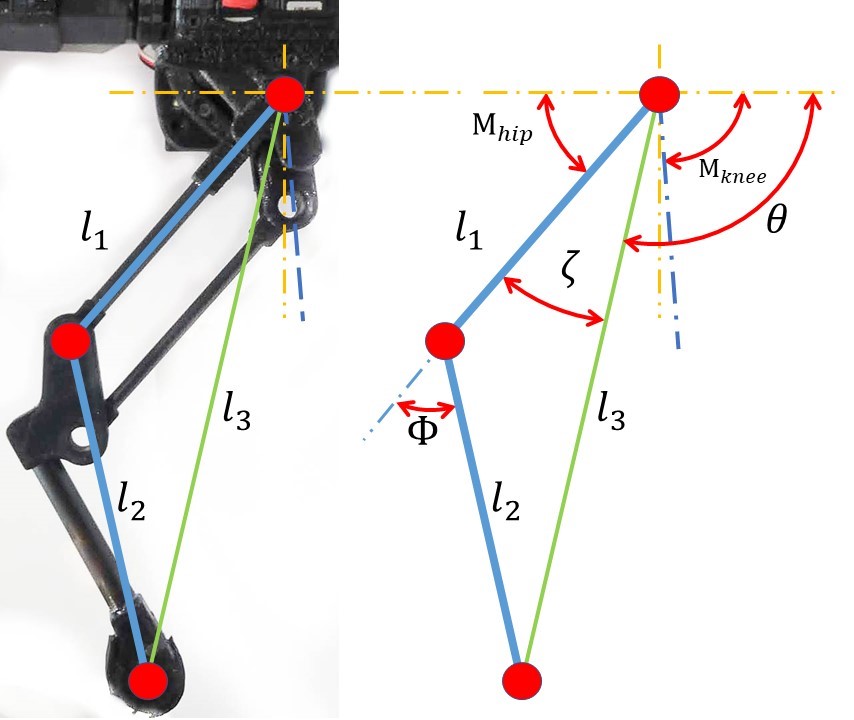}
\caption{Inverse kinematics of robot leg.}
\label{fig:Inv_kinematics}
\end{figure}

Having obtained the coordinates (either in the form of polar coordinates ($r,\theta$) or Cartesian coordinates ($x,y$) of end points), we generate the joint angles via inverse kinematics. \Cref{fig:Inv_kinematics} describes the leg configuration and the variables.
If the end points of the robot are described in Cartesian coordinates, then \Crefrange{eq:IK_1}{eq:IK_4} are used to determine the joint angle.
In case the description is in polar co-ordinate, \Cref{eq:IK_1} can be avoided, since $l_3$ and $\theta$ are directly provided.
The inverse kinematics are given by 
\begin{align}
    & l_3 = \sqrt{x^2 + y^2},~~\theta =  \arctan\left(\dfrac{y}{x}\right)\label{eq:IK_1}\\
    & \Phi = \arccos \left( \dfrac{l_3^2 - l_1^2 - l_2^2}{2 l_1 l_2}\right)\label{eq:IK_2}\\
    & \zeta = \arctan \left(\dfrac{l_2 \sin{\Phi}}{l_1 + l_2 \cos{\Phi}}\right) \label{eq:IK_3}\\
    & q_{hip} = \theta + \zeta,~~q_{knee} = \Phi + q_{hip}, \label{eq:IK_4}
\end{align}
where, $ l_1,l_2 $ are the hip and knee lengths of the robot.
Since each robot has a different reference frame associated with legs, the joint angles are linearly transformed to obtain motor angles.
The motor signals are then transformed to a pulse width modulation (PWM) signal to drive the motors.

\section{Experimental Results}\label{sec:results}


In this section, we will describe various strategies undertaken to generate the different types of robot behaviors. 
First, we discuss the various types of flat ground walking gaits. Next, we describe the abduction free turning, 
and then finally describe the transitions between these gaits.

\subsubsection{Flat ground walking}
\label{FlatPlaneMotion}
On the flat plane, we have introduced $6$ different types of gaits, trot, gallop, bound, walk, modified trot $1$ and modified trot $2$.
We have used the basis function based approach \cite{singla2018realizing} to store the information about the trajectories in a few constant values. These constant values are stored in the weight matrix ($W_x, W_y$). 
Using the weights, we can obtain the desired end point trajectories and the velocities. We get
\begin{align}
    x_{i}= \sum_{j=0}^{5}f_{k,j}(\phi_i) W_{x,i,j}, ~~y_{i}= \sum_{j=0}^{5}f_{k,j}(\phi_i) W_{y,i,j},
\end{align}
where, $x_{i}, y_{i}$ are the target x, y coordinates of the $i^{th}$ leg. $f_{k,j}(\phi_i)$ is the $j^{th}$ kinematic motion primitive \cite{singla2018realizing}, which essentially is a polynomial of $\phi_i$. The comparison of speeds obtained for various gaits. A more detailed description of the basis functions are given in \cite{singla2018realizing}.


\subsubsection{Turning}
\label{Turning}
The current model of Stoch, does not contain any abduction. This limits us to realization of abduction free/locked turning. We set the $X$ axis (forward-backward) movement of the desired leg to zero, and use 
\begin{align}
& x_{turn,i} = \alpha_{t,i} x_{i}\\
& \alpha_{t} = 
  \begin{cases}
    \left[1, 0, 0, 1\right] & \quad \text{left turn}\\
    \left[0, 1, 1, 0\right] & \quad \text{right turn}\\
    \left[1, 1, 1, 1\right] & \quad \text{no turn} 
  \end{cases}
\end{align}


where $\alpha_{t}$ is the vector containing the turning coefficients and $x_{turn,i}$ is the target $x$ axis position of the $i^{th}$ leg. To avoid discontinuous changes in the motion, we pass the $\alpha_{t}$ through a low pass filter, \Cref{eq:lowpassfilter}. This results in a smooth trot-turn transition. 

\subsubsection{Gait transition}
\label{Transition}
In this section, we will provide a general methodology for transitioning between the flat-ground walking gaits.
In \Cref{sec:PatternGenerator}, we described the phase offset, $\Phi$, as a constant set of values which can be changed to obtain various gaits. This change can be achieved, on the fly, by passing the desired phase offset through a low pass filter:
\begin{align}
\dot{\Phi}_i = \alpha_{\Phi}(\Phi_{d,i} - \Phi)
\end{align}
where $\Phi_{d,i}$ is the desired phase offset of the $i^{th}$ leg, and $\Phi_i$ is the current phase offset of the $i^{th}$ leg. This allows for a smooth change in the phase values. 

\subsection{Video Result}

Video results of walking experiments on the \textit{Stoch} are available on \url{https://youtu.be/Wxx9pwwTIL4}.
Specifically, we show trot, bound, gallop, walk gaits on the robot.
Additionally turning and gait transitions are also shown. 
Further, the animation and physical assembly is also shown in the video for easy understating of the \textit{Stoch} design.

\section{CONCLUSION} \label{sec: conclusion}

This work presents a custom built quadruped robot along with its design, software, control framework and experimental validation on the hardware. 
Compared to other existing quadruped robots, our platform requires less resource and costs under \$$1000$.
Several gaits were realized such as walk, trot, gallop, and bound. Additionally, the robot can turn with a small radius via CPG based gait transitions, without the use of abduction. The robot reached a maximum forward speed of $0.6$ m/s. An open-loop controller was used to set the speed and direction in all these experiments.
Future work involves increasing the payload, and also incorporate  external sensors for closed-loop control of the quadruped. 





\section*{ACKNOWLEDGMENT}
We acknowledge Ashish Joglekar, Rokesh Laishram and Balachandra Hegde for help in software development, Carbon fibre component manufacturing and PCB design.

\medskip

\bibliographystyle{IEEEtran}
\bibliography{references}

\end{document}